\RequirePackage{fix-cm}
\documentclass{article}
\PassOptionsToPackage{table}{xcolor}
\usepackage{iclr2027_conference,times}
\usepackage{amsmath,amssymb}
\usepackage{booktabs}
\usepackage{graphicx}     
\usepackage{marvosym}
\usepackage{textcomp}
\usepackage{array}
\usepackage{multirow}
\usepackage{xcolor}
\usepackage{mdframed} \usepackage{fvextra}
\usepackage{xcolor}
\usepackage[colorlinks=true, linkcolor=blue, urlcolor=blue]{hyperref}
\usepackage{hyperref}
\usepackage{url}
\usepackage{xurl}
\usepackage{pifont}   
\hypersetup{hidelinks}

\definecolor{videoBlue}{RGB}{43,108,176}
\definecolor{bridgeGreen}{RGB}{43,122,96}
\definecolor{webRed}{RGB}{177,69,58}
\definecolor{figGreen}{HTML}{00B050} 
\newcommand{\method}{\textsc{GroundedVDR}}

\definecolor{checkgreen}{RGB}{18,145,63}
\definecolor{crossred}{RGB}{230,28,28}
\definecolor{partialorange}{RGB}{205,124,0}

\definecolor{visionheader}{RGB}{234,243,250}
\definecolor{visionrow}{RGB}{247,251,254}
\definecolor{videoheader}{RGB}{240,238,246}
\definecolor{videorow}{RGB}{251,250,253}
\definecolor{grokheader}{RGB}{253,243,232}
\definecolor{grokrow}{RGB}{254,250,246}
\definecolor{qwenheader}{RGB}{253,238,238}
\definecolor{qwenrow}{RGB}{254,248,248}
\definecolor{oursrow}{RGB}{238,247,233}

\newcommand{\cmark}{\textcolor{checkgreen}{\ding{51}}}
\newcommand{\xmark}{\textcolor{crossred}{\ding{55}}}
\newcommand{\pmark}{\textcolor{partialorange}{\raisebox{0.10ex}{\scriptsize$\circ$}}}

\title{GroundedVDR: Grounding Video-to-Web Evidence Dependencies in Video Deep Research}

\DeclareRobustCommand{\envmark}{\begingroup
  \sbox0{1}\sbox2{\scalebox{0.6}{\Letter}}%
  \raisebox{\dimexpr(\ht0-\ht2+\dp2)/2\relax}{\usebox2}\endgroup}

\author{\mdseries%
    \textbf{Xiyuan Liang}\textsuperscript{1,$\ast$}
    \quad
    \textbf{Yu Wang}\textsuperscript{1,$\ast$}
    \quad
    \textbf{Zuhao Yang}\textsuperscript{1,\textdagger}
    \quad
    \textbf{Ziting Wang}\textsuperscript{1}
    \quad
    \textbf{Duo Li}\textsuperscript{1}
    \\
    \textbf{Sudong Wang}\textsuperscript{2}
    \quad
    \textbf{Zhongyu Yang}\textsuperscript{3}
    \quad
    \textbf{Xiaoqin Zhang}\textsuperscript{4}
    \quad
    \textbf{Ling Shao}\textsuperscript{5}
    \quad
    \textbf{Shijian Lu}\textsuperscript{1,\envmark}
  \\ \textsuperscript{1}NTU \quad \textsuperscript{2}HKUST(GZ) \quad \textsuperscript{3}ModelBest \quad \textsuperscript{4}ZJUT \quad \textsuperscript{5}UCAS
  \\
  \texttt{\{WANG2361,XIYUAN006\}@e.ntu.edu.sg}
  \\
  \textsuperscript{$\ast$}Equal Contribution.
  \qquad
  \textsuperscript{\textdagger}Project Lead.
  \qquad
  \textsuperscript{\envmark}Corresponding Author.
  }

\newif\ifpreprint
\preprinttrue
\ifpreprint\iclrfinalcopy\fi

\begin{document}

\maketitle
\ifpreprint\lhead{Preprint}\fi
\begin{abstract}
Video large multimodal models have improved at identifying cross-frame visual evidence and conducting open-web retrieval, as reflected in recent video deep research (VDR) benchmarks. However, existing VDR benchmarks primarily evaluate tool-use trajectories and final responses without verifying whether a model actually uses the temporal video information or the retrieved web evidence. To fill this gap, we introduce \method{}, an evidence-grounded benchmark with three features. First, it combines human review with restricted-access probing across multiple models to exclude questions answerable from parametric knowledge or a single information source. Second, each question carries structured video-to-web evidence annotations that link the relevant video segments to their supporting web evidence. Third, its evaluation protocol separately scores video grounding, web evidence grounding, and final-answer correctness. Across eight multimodal agent systems, the best reaches 76.0\% answer accuracy, yet none recovers the complete required web evidence for more than 8\% of questions. Our benchmark is publicly available at \url{https://huggingface.co/datasets/hh721/vdr_temp}.

\end{abstract}

\section{Introduction}

Most existing video understanding benchmarks already evaluate a range of video reasoning abilities, but the reasoning they require usually remains within the video itself \citep{li2024mvbench,wu2024longvideobench,fu2025video,zhou2025mlvu,fang2024mmbench}. On the other hand, many text-based deep-research studies examine how models perform multi-hop search and reasoning over the open web from text queries \citep{wei2025browsecomp,wu2025webwalker,li2025webthinker,zheng2025deepresearcher}. In practice, however, people often need to combine video understanding with web search to study complex questions, such as those arising in film analysis, sports-event analysis, or knowledge exploration. Video deep research (VDR) was introduced to address this need by requiring evidence from the video to determine what should be researched on the web and how the search should proceed.

Recent studies have begun to develop agent-based methods and benchmarks for VDR. VideoDR \citep{liu2026watching} formalizes the task and introduces modality-dependence filtering during dataset construction. VideoSearcher \citep{gao2026videosearcher} further evaluates fine-grained video interaction and search tools, while Video-DeepResearch \citep{fang2026videodeepresearch} studies common failures such as modality bias and parametric knowledge leakage. Together, these works establish the basic VDR setting and extend its evaluation beyond final-answer accuracy.

\begin{figure}[!t]
  \centering
  \includegraphics[width=1\linewidth]{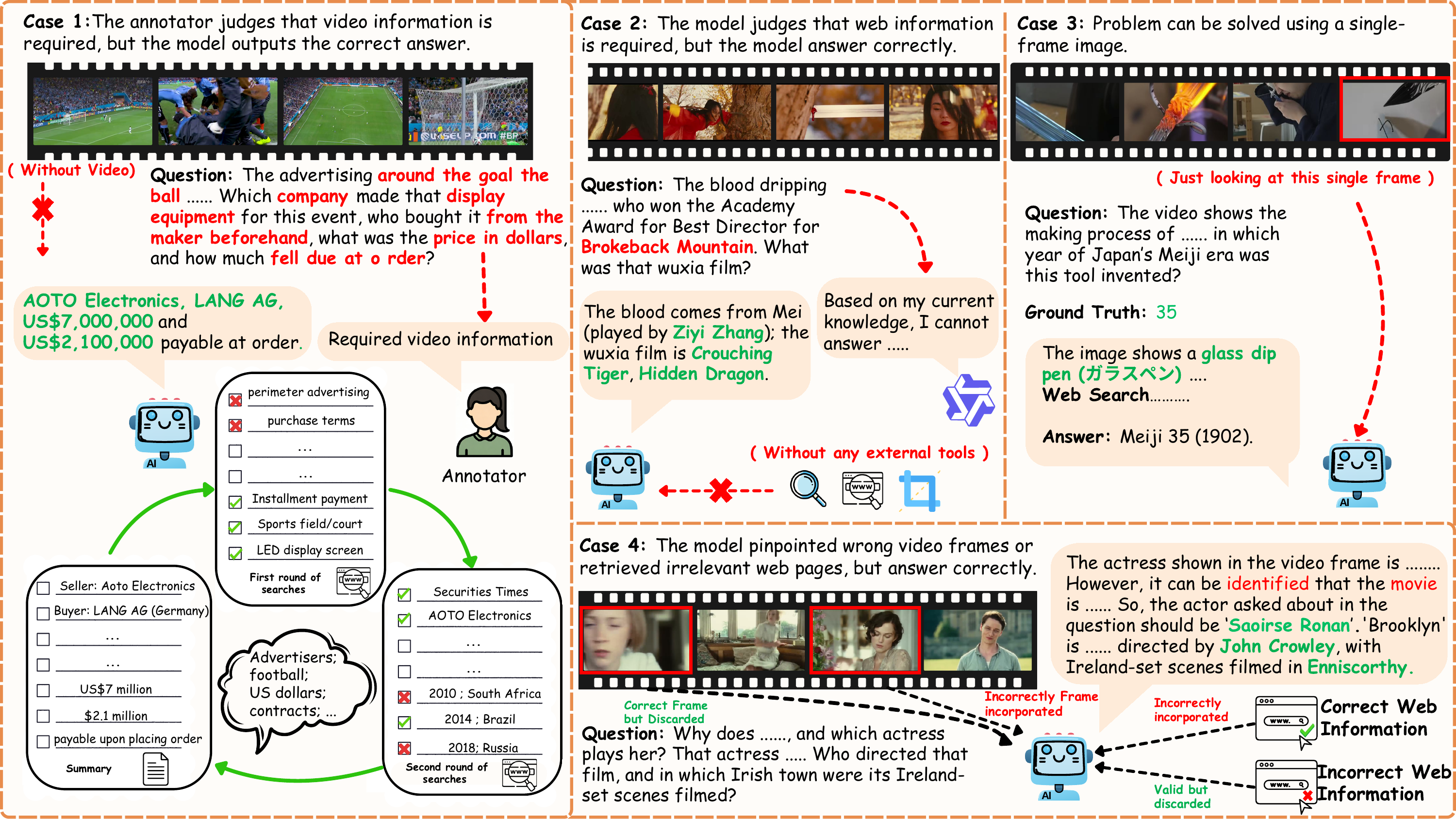}
    \caption{\textbf{Failure cases in video deep research (VDR).} Case 1 bypasses the video, and Case 2 bypasses web retrieval, illustrating variation in human and model judgments of modality dependence. Case 3 reduces the task to single-frame entity recognition rather than video-guided search. Case 4 produces the correct final answer from an incorrect video segment and irrelevant web evidence. \textcolor{figGreen}{Green text} marks correct answers and correctly identified entities, while \textcolor{red}{red text} indicates information leakage.}
  \label{fig:motivation}
\end{figure}
However, many questions in current VDR benchmarks can still be solved using video-side clues obtained from a set of static frames and on-screen text, such as signs or subtitles. Such questions make limited use of richer temporal information and dynamic changes of entities in the video. Their temporal reasoning is also often limited to temporal order, cross-frame matching, or simple comparisons between entities, rather than more complex relations such as causality, state changes, and evidence accumulation over time. In addition, current benchmarks provide limited evaluation of omnimodal video understanding, such as using audio information or jointly reasoning over visual and audio evidence. 
Evaluating VDR systems also remains difficult because conventional metrics do not fully capture the research process. Final-answer accuracy cannot distinguish whether a model reaches the answer from the correct evidence or simply recalls it from parametric knowledge.  Tool-use statistics only record which actions were performed, but do not show whether those actions were relevant or whether the retrieved evidence was correct. 

Specifically, existing VDR benchmarks face three related challenges, as shown in Figure~\ref{fig:motivation}.
The \textbf{first} challenge is deciding whether a question requires both video and web evidence. Existing benchmarks typically rely on either human annotators \citep{liu2026watching,gao2026videosearcher} or model-assisted filtering and judging \citep{fang2026videodeepresearch}. 
However, these checks depend on the prior knowledge and capability of the evaluator. As shown in Figure~\ref{fig:motivation}, a model might bypass the intended video (Case 1) or web retrieval (Case 2) entirely and still obtain the correct answer. 
The \textbf{second} challenge is ensuring that the video constrains the web search rather than merely providing a searchable keyword. In some benchmarks, recognizing a single visual entity is enough to find the answer (Case 3).
The \textbf{third} challenge is verifying the validity of the research process. An agent might locate the wrong video segments, retrieve irrelevant webpages, and misinterpret the evidence, yet still guess the correct final answer (Case 4). Final-answer accuracy and tool-use statistics cannot prove whether the agent actually relies on the correct evidence. In addition, they fail to reveal where the reasoning chain breaks down.

To address these challenges, we introduce \method{}, an evidence-grounded benchmark that evaluates the video-to-web dependency. 

First, we combine restricted-access probing with human review. We probe each question under question-only, video-only, and web-only conditions, with one or both information sources withheld. A question is retained only when joint video--web evidence determines the answer, while each restricted condition leaves it underdetermined. 
The no-web conditions are cross-checked across several vision-language models, so shortcuts available to different models can be detected, reducing dependence on a single tester.
Second, we organize the required video evidence into semantic video states, each summarizing a video-side fact supported by one or more temporal segments. Each annotation also records the constraints derived from these states, the supporting web evidence, and evidence-backed incorrect candidates. Third, our evaluation protocol scores video grounding, web evidence grounding, and final-answer correctness separately, so an error can be traced to the stage where it occurs.

In this work, our contributions are fourfold:
\begin{itemize}
    \item We rethink VDR benchmark construction as a necessity-testing problem. Human review and multi-model probing require each task to depend on both temporal video grounding and web retrieval, reducing parametric-memory shortcuts.
    \item We introduce \method{}, a benchmark with structured evidence annotations that link video evidence to web search constraints and supporting web claims.
    \item We develop a stage-wise evaluation protocol that separately scores video grounding, web evidence grounding, and final-answer correctness, allowing us to identify where errors occur.
    \item We evaluate eight multimodal agent systems on \method{} and find that correct answers often come with incomplete supporting evidence. We also identify failures in carrying video-derived constraints through later web research as a major challenge.
\end{itemize}

\section{Related Work}
\begin{table}[t]
\centering
\caption{\textbf{Comparison of benchmark construction, necessity tests, and evidence annotations.}
\textbf{Seed Direction} gives the order in which the task is constructed, and
\textbf{Annotation (Anno.)}\ identifies manual or hybrid construction.
\textbf{Video-only/web-only tests (V/W Tests)} check whether either modality alone is sufficient.
\textbf{Multi-model} indicates testing across multiple models without search access.
\textbf{Multi-hop} and \textbf{Multi-state} indicate multiple reasoning steps and temporally distinct video states, respectively.
\textbf{Video-to-web links (V$\rightarrow$W Links)} connect localized video evidence to web retrieval conditions.
\textbf{Claim-level evidence (Claim Evid.)}\ records supporting text from web sources, and \textbf{hard negatives (HN)} are evidence-backed incorrect candidates.
\cmark, \pmark, and \xmark{} denote full, partial, and absent support; N/A denotes not applicable.}
\label{tab:benchmark-construction-comparison}

\setlength{\tabcolsep}{3.2pt}
\renewcommand{\arraystretch}{1.08}

\resizebox{\textwidth}{!}{%
\begin{tabular}{@{}lccccccccc@{}}
\toprule
& \multicolumn{2}{c}{\textbf{Basic Construction}}
& \multicolumn{4}{c}{\textbf{Necessity Tests}}
& \multicolumn{3}{c}{\textbf{Evidence Annotations}} \\
\cmidrule(lr){2-3}
\cmidrule(lr){4-7}
\cmidrule(lr){8-10}
\textbf{Benchmark}
& \textbf{Seed Direction}
& \textbf{Anno.}
& \textbf{V/W Tests}
& \textbf{Multi-model}
& \textbf{Multi-hop}
& \textbf{Multi-state}
& \textbf{V$\rightarrow$W Links}
& \textbf{Claim Evid.}
& \textbf{HN} \\
\midrule

\rowcolor{visionheader}
\multicolumn{10}{@{}l}{\textbf{\textit{Image Deep-Research Benchmarks}}} \\

\rowcolor{visionrow}
BrowseComp-VL \citep{geng2025webwatcher}
& Text$\rightarrow$Image & Hybrid
& \xmark & \xmark & \cmark & N/A
& \pmark & \xmark & \xmark \\

\rowcolor{visionrow}
MMSearch$^{\star}$ \citep{jiang2024mmsearch}
& Query-first & Manual
& \xmark & \cmark & \pmark & N/A
& \pmark & \pmark & \xmark \\

\rowcolor{visionrow}
VDR-Bench \citep{zeng2026visiondeepresearchbenchmark}
& Image-first & Hybrid
& \pmark & \pmark & \cmark & N/A
& \pmark & \pmark & \xmark \\

\midrule
\rowcolor{videoheader}
\multicolumn{10}{@{}l}{\textbf{\textit{Video Deep-Research Benchmarks}}} \\

\rowcolor{videorow}
VideoDR \citep{liu2026watching}
& Video-first & Manual
& \cmark & \xmark & \cmark & \cmark
& \pmark & \pmark & \xmark \\

\rowcolor{videorow}
VideoSearcher \citep{gao2026videosearcher}
& Video-first & Manual
& \cmark & \xmark & \cmark & \pmark
& \pmark & \pmark & \xmark \\

\rowcolor{videorow}
Video-DeepResearch \citep{fang2026videodeepresearch}
& Video-first & Hybrid
& \pmark & \pmark & \cmark & \pmark
& \pmark & \pmark & \xmark \\

\rowcolor{videorow}
VideoRover-Bench \citep{liu2026videorover}
& Target$\rightarrow$Video & Hybrid
& \pmark & \pmark & \pmark & \pmark
& \pmark & \pmark & \xmark \\

\midrule
\rowcolor{oursrow}
\textbf{\method{} (Ours)}
& \textbf{Video/Evidence-first}
& \textbf{Manual}
& \cmark & \cmark & \cmark & \cmark
& \cmark & \cmark & \cmark \\
\bottomrule
\end{tabular}%
}

\begin{minipage}{\textwidth}
\footnotesize
$^{\star}$The full MMSearch benchmark is mixed, containing 171 image-query instances and 129 text-only instances. The comparison here concerns its image-query setting.
\end{minipage}

\end{table}
\paragraph{Video Understanding.}
Existing video benchmarks primarily evaluate a wide range of perception and temporal reasoning \citep{li2024mvbench,wu2024longvideobench,zhou2025mlvu,wang2025lvbench}. Video-LLMs have also been increasingly studied for fine-grained temporal grounding \citep{yang2025timeexpert}, while agent-based methods allow models to actively search long videos through adaptive sampling and tool \citep{zhang2025deep,yang2026longvt,yang2026paravt,chen2025lvagent,yang2026svagent}. Notably, LongVidSearch advances this line of work by enforcing retrieval necessity through discontinuous evidence clips \citep{yu2026longvidsearch}. However, a fundamental limitation persists across these paradigms: they all assume the necessary information resides entirely within the video itself. In contrast, our benchmark requires agents to extract temporal video states and use them as strict constraints for subsequent open-web research.

\paragraph{Deep-Research Agents.}
Text-based deep-research systems search the web and combine information from multiple sources \citep{wei2025browsecomp,wu2025webwalker,li2025webthinker,zheng2025deepresearcher}. Recent studies extend this setting to multimodal search with visual inputs and multiple retrieval tools \citep{jiang2024mmsearch,geng2025webwatcher}. ALCE evaluates whether retrieved sources support generated claims \citep{gao2023alce}, while the Vision-DeepResearch Benchmark (VDR-Bench) shows that visual-search evaluations can contain text-only shortcuts and parametric knowledge leakage \citep{zeng2026visiondeepresearchbenchmark}. These multimodal studies focus on static images; we derive the search constraints from temporally distributed video evidence.

\paragraph{Video Deep Research.}
Recent studies begin to connect video understanding with open-web research. VideoDR defines the video deep research (VDR) task and uses human filtering to ensure dependence on both video and web information \citep{liu2026watching}. Video-BrowseComp emphasizes that open-web research should depend on temporal video evidence \citep{liang2025videobrowser}. VideoSearcher focuses on agent tool use, combining temporal localization, spatial inspection, and multimodal search \citep{gao2026videosearcher}; Video-DeepResearch uses a perception-before-retrieval procedure to reduce modality bias \citep{fang2026videodeepresearch}. These studies cover multi-frame reasoning, open-web retrieval, and agent tool use. As shown in Table~\ref{tab:benchmark-construction-comparison}, \method{} additionally annotates the links among video evidence, search constraints, and claim-level web evidence, and tests whether each link is necessary.

\section{GroundedVDR}
\label{sec:task}
\subsection{Preliminaries}
\label{sec:task-and-evidence}

\paragraph{Task Definition.}
We formulate video deep research (VDR) as a cross-modal \emph{information-completion} task. An instance provides a video $V$, a question $q$, and access to the open web $W$. The question specifies a required information set $R(q)$: the minimal facts needed to determine the final answer $y$, which may consist of several answer fields $y=(y_1,\ldots,y_F)$. This set comprises a video-grounded component $R_V(q)$ and a web-derived component $R_W(q)$. A complete solution must establish and synthesize both components. Consequently, a VDR system outputs an answer alongside its supporting evidence, reported as a tuple $(\hat{y}, \hat{\mathcal{E}}_V, \hat{\mathcal{E}}_W)$, where the submitted video evidence $\hat{\mathcal{E}}_V$ and web evidence $\hat{\mathcal{E}}_W$ must sufficiently ground the required components $R_V(q)$ and $R_W(q)$.

\paragraph{Evidence Representation.}
To evaluate whether a system resolves $R(q)$, we record the required evidence as structured annotations (Appendix~\ref{app:annotation}). First, the video requirement $R_V(q)$ is formalized as a set of required video states \(\mathcal{S}^{\star}_q = \{s_1,\ldots,s_M\}\). Each state \(s_j\) is a semantic sub-fact supported solely by the video and grounded to one or more minimal temporal evidence units \(\mathcal{U}(s_j)=\{u_{j,1},\ldots,u_{j,n_j}\}\). Each unit records a time interval, its modality (visual, audio, or audiovisual), and the observed content. This separates semantics from localization: evidence units identify \emph{where} the information occurs, while states specify \emph{what} video-side fact is established. Second, the web requirement $R_W(q)$ is formalized as a set of required semantic claims \(\mathcal{C}^{W}_q = \{c^W_1,\ldots,c^W_K\}\). Each claim captures a necessary external fact and is annotated with at least one supporting web source; other sources that establish the same fact are also accepted.
Finally, video-to-web links specify how the required video states determine which external claims must be established. We also annotate \emph{candidate-linked hard negatives}: plausible but incorrect answer candidates supported by relevant web evidence but inconsistent with the full video-to-web constraints of the question.

\subsection{Benchmark Construction}
\label{sec:construction}
\begin{figure}[!t]
  \centering
  \makebox[\linewidth][c]{%
    \includegraphics[
      width=1.05\linewidth,
      clip
    ]{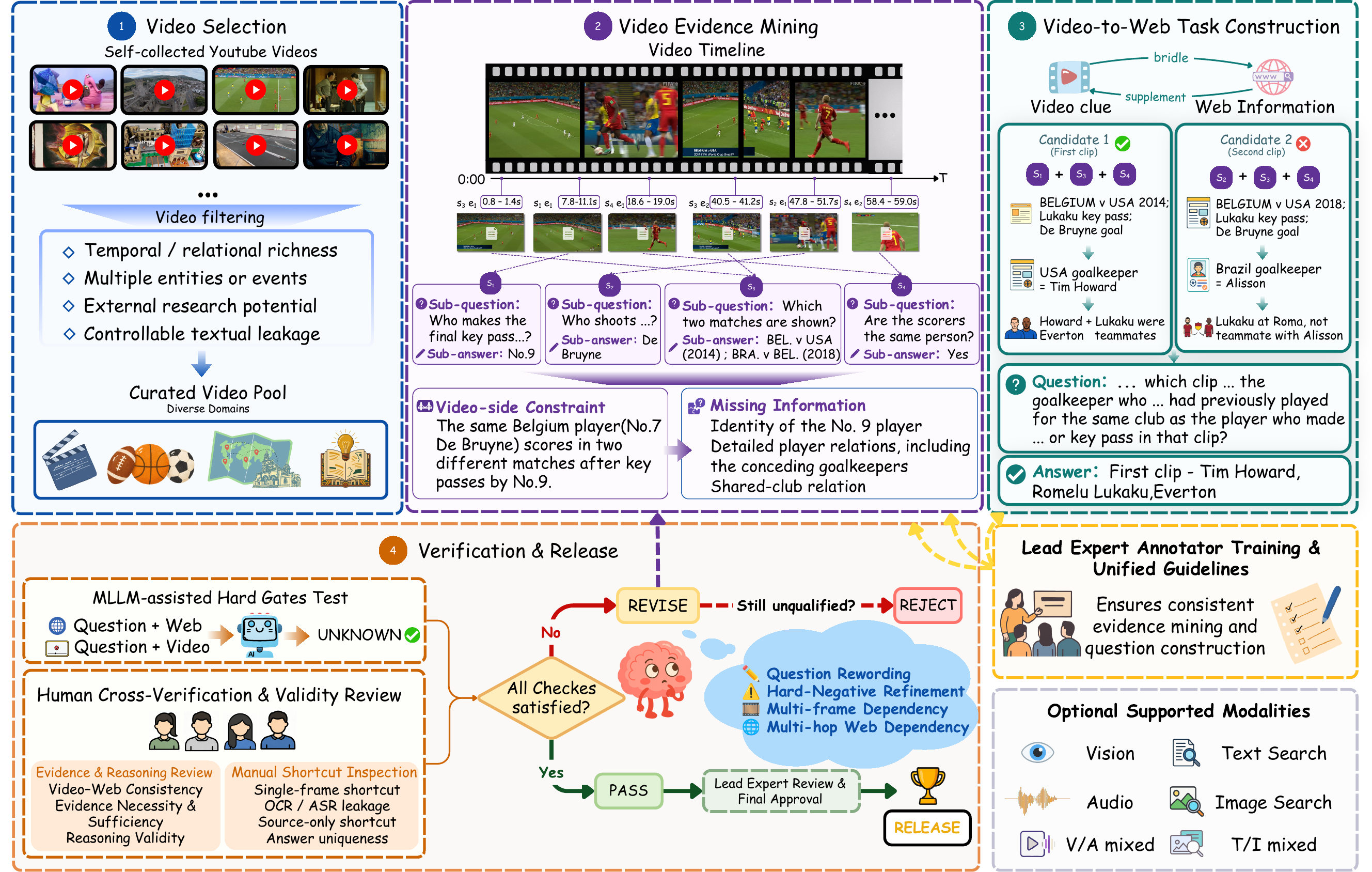}%
  }
\caption{\textbf{Four-stage construction procedure for \method{}.}
The stages are (1) video selection, (2) video evidence mining, (3) video-to-web task curation, and (4) verification and release. Questions are retained only when the annotated video evidence constrains the subsequent web search and the supporting web claims pass automated checks and human review.}
  \label{fig:framework}
\end{figure}

We construct \method{} through the four-stage pipeline in
Figure~\ref{fig:framework}. \textbf{Video Selection} first keeps videos whose evidence is distributed across multiple time segments, involves multiple entities or events, and can be connected to reliable external sources. It removes videos that show a single scene or whose on-screen text directly reveals key information. It also removes videos that mainly support simple factual questions or whose relevant facts cannot be checked on the web (Figure~\ref{fig:framework}(1)). \textbf{Video Evidence Mining} then inspects
the timeline and aggregates minimal observations across multiple moments into
video-side states and constraints. Evidence is collected before any question is written, so each question is built from the evidence (Figure~\ref{fig:framework}(2)). 
\textbf{Video-to-Web Task Curation}
uses these video constraints to guide web research and to write the question.
Annotators collect claim-level supporting evidence and plausible alternative
candidates, including candidate-linked hard negatives when available
(Figure~\ref{fig:framework}(3)). Finally, \textbf{Verification and Release}
checks temporal necessity, video-to-web dependence, evidence validity, and
shortcut resistance before an instance enters the benchmark
(Figure~\ref{fig:framework}(4)).

\subsection{Quality Assurance}
\label{sec:quality-control}
Every question passes two checks before release: multi-model probing and independent human review. An instance is valid only when the joint video--web evidence determines the answer and neither modality alone is sufficient. This verification-and-release stage is
summarized in Figure~\ref{fig:framework}(4) and implemented through the checks below.

\paragraph{Restricted-Access Probing.}
We probe each candidate question under three restricted conditions to detect shortcuts. The first two withhold web access: \emph{question-only} (no video, no web) and \emph{video-only} (video, no web). They test whether the answer is reachable from parametric knowledge, with or without the video. We run both with several vision-language models that from different model families, so a shortcut available to one model's prior is still caught by cross-checking across models. The third condition, \emph{web-only} (question and web search, no video), tests whether web research alone determines the answer, and we run it with a closed-source web-search agent with multi-step browsing and open-web retrieval capabilities. In every condition we draw multiple samples rather than a single greedy output, since one correct sample is enough to reveal a shortcut. A fixed large language model (LLM) judge compares each sample with the ground-truth answer and its aliases, and any question solved under a restricted condition is sent to human review.

\paragraph{Independent Human Cross-Verification.}
All candidate questions, including those flagged by probing, are reviewed by annotators who did not write them. For a flagged question, reviewers read the model output and decide why it succeeded. The question is removed if the success comes from parametric knowledge, or from reading the answer directly through optical character recognition (OCR) or automatic speech recognition (ASR),since either bypasses the intended joint video--web reasoning. It is kept if the success is a lucky guess or a judge error. For every question, reviewers also check that each required state is necessary, each cited source entails its claim, the incorrect candidates are plausible to a human reader, and no single webpage answers the question alone. A third expert resolves disagreements. Inter-annotator agreement is 0.76 (Cohen's $\kappa$), and 24\% of candidate questions are removed at this stage.

\begin{figure*}[!t]
    \centering
    \includegraphics[width=0.82\textwidth]{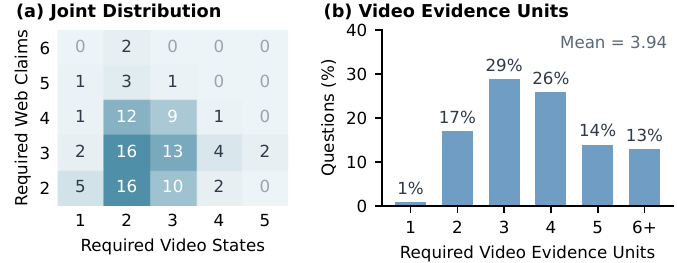}
    \caption{\textbf{Evidence structure of \method{}.}
    (a) Joint distribution of the numbers of required video states and web
    claims; each cell reports the number of questions.
    (b) Distribution of temporally localized video evidence units required per
    question.}
    \label{fig:benchmark-profile}
\end{figure*}

\begin{figure*}[!t]
    \centering
    \includegraphics[width=\textwidth]{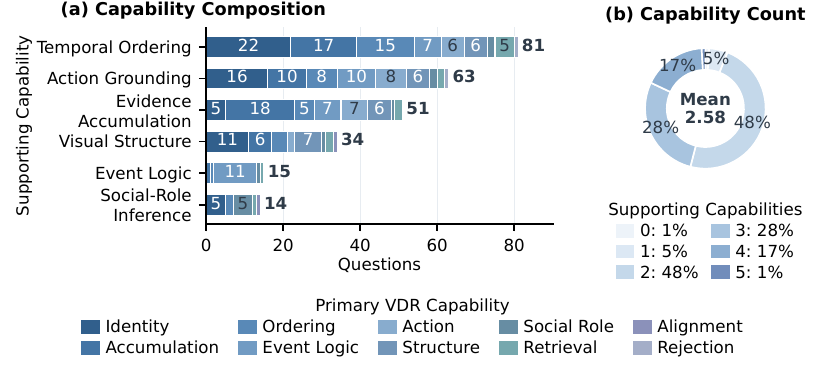}
\caption{\textbf{Video capabilities required by \method{}.}
(a) Supporting capabilities associated with each primary video deep research (VDR) capability.
(b) Number of supporting video capabilities jointly required by each question.}
    \label{fig:capability-profile}
\end{figure*}

\paragraph{Benchmark Profile.}
\label{sec:benchmark-profile}
Figures~\ref{fig:benchmark-profile} and~\ref{fig:capability-profile}
characterize the structure and video-side capabilities of the released
benchmark. Figure~\ref{fig:benchmark-profile} shows that most questions require
multiple video states, multiple web claims, and evidence from multiple temporal
segments, rather than a single visual lookup followed by direct retrieval. Figure~\ref{fig:capability-profile} provides a complementary view of the video-side reasoning requirements. It shows that the benchmark does not isolate individual abilities: many questions require several temporal, spatial, and relational capabilities to work together. This suggests that successful VDR depends not only on finding the right video evidence, but also on combining different types of video reasoning before that evidence can guide web research.

\subsection{Evaluation Metrics and Protocol}
\label{sec:evaluation_protocol}
\paragraph{Auditing the Research Process.}
We evaluate the submitted evidence against the structured video states and web
claims defined in Section~\ref{sec:task-and-evidence}. For each question,
\textbf{Video State} measures the proportion of its required video states that
are supported by the submitted video evidence, while \textbf{Web Claim}
measures the proportion of its required web claims that are supported by the
submitted web evidence. Both metrics are computed per question and then
macro-averaged across the benchmark, so that every question contributes
equally regardless of how many states or claims it contains. For further analysis, we also
characterize question-level grounding as \textbf{Complete}, \textbf{Partial},
or \textbf{None}, depending on whether all, some but not all, or none of the
required states or claims are supported.

\paragraph{Outcome Evaluation.}
We separately evaluate whether the final answer resolves the information
requested by the question. For each question, \textbf{Field Acc.}\ measures the
proportion of its required answer fields that are correctly resolved, and is
likewise macro-averaged across questions. \textbf{Answer} is the stricter
question-level measure: a question is counted as correct only when all of its
required answer fields are correctly resolved. Field Acc.\ therefore gives
partial credit to multi-field answers with only some fields correct, whereas Answer measures
complete answer resolution.

\paragraph{Semantic Judgment.}
For answer fields, video states, and web claims, we use a fixed LLM judge (GPT-5.6 Terra) with task-specific rubrics. The judge evaluates semantic support rather than requiring exact string matches
or exact agreement with the annotated temporal boundaries. For video evidence, submitted temporal segments need not match the annotated
time boundaries exactly, as long as they provide sufficient support for
the required video state. For web evidence, alternative reliable sources and paraphrased text are accepted when they support the required claim; merely related pages and evidence attached to an incorrect candidate are rejected.

\section{Experiments}
\label{sec:experiments}
\subsection{Implementation Details}
\label{sec:eval-setting}

\paragraph{Platforms.}
We evaluated each model through an agent platform that supports video inspection,
web browsing, and tool use. Each system received the same video,
question, and task instructions. It returned a final answer and the supporting video and web evidence in a common format.
We used each model's native platform when available. Because platforms may differ in tools and orchestration, we treat each model--platform pair as an evaluated system. The results should not be interpreted as isolating model-level effects.

\paragraph{Models.}
We evaluated eight multimodal agent systems: Claude Opus~4.8~\citep{anthropic2026claude48} and Claude Sonnet~5~\citep{anthropic2026sonnet5};
GPT-5.6 Sol and GPT-5.6 Terra~\citep{openai2026gpt56};
Grok~4.6~\citep{xai2026grok4} and Composer~2.5~\citep{cursor2026composer};
and Qwen3.8-Flash~\citep{qwen2026qwen3} and Qwen3.8-Max~\citep{qwen2026qwen38}.
All systems were evaluated on the same released 100-question benchmark. We report the agent platform and reasoning setting for each system in Appendix~\ref{app:model-platform-details}.
\paragraph{Human-Expert Baseline.}
We evaluated three human experts under the same task definition. They received the video, question, and open-web access, and submitted a final answer with supporting video and web evidence. They could use standard web browsers, translation services, basic calculators, and manual video controls. They could not use conversational language models, AI agents, browser assistants, or automated research tools. 

\subsection{Main Results}
\label{sec:main-results}

\begin{table}[t]
\centering
\caption{\textbf{Main results.}
Answer metrics evaluate the final response, whereas video and web metrics
measure recovery of the required semantic evidence. \textbf{Field Acc.} is the
mean per-question proportion of correctly resolved required answer fields,
while \textbf{Answer} is the proportion of questions whose required answer fields are all
correct. \textbf{Video State} and \textbf{Web Claim} are the mean per-question
recalls of required video states and web claims, respectively. Best agent results in each column are shown in bold; the human-expert baseline
is reported separately for reference.}
\label{tab:main-results}
\setlength{\tabcolsep}{3.6pt}
\renewcommand{\arraystretch}{1.08}
\resizebox{0.85\linewidth}{!}{%
\begin{tabular}{@{}lrrrr@{}}
\toprule
\textbf{System}
& \textbf{Field Acc.}
& \textbf{Answer}
& \textbf{Video State}
& \textbf{Web Claim} \\
\midrule

\rowcolor{visionrow}
\multicolumn{5}{@{}l}{\textbf{\textit{Claude Series}}} \\
\rowcolor{visionrow}
Claude Opus 4.8
& 61.3 & 55.0 & \textbf{43.5} & 16.9 \\
\rowcolor{visionrow}
Claude Sonnet 5
& 61.6 & 55.0 & 31.2 & 17.9 \\

\midrule
\rowcolor{videorow}
\multicolumn{5}{@{}l}{\textbf{\textit{GPT Series}}} \\
\rowcolor{videorow}
GPT-5.6 Sol
& \textbf{82.0} & \textbf{76.0} & 37.6 & \textbf{32.0} \\
\rowcolor{videorow}
GPT-5.6 Terra
& 55.5 & 47.0 & 20.5 & 10.6 \\

\midrule
\rowcolor{grokrow}
\multicolumn{5}{@{}l}{\textbf{\textit{Grok / Composer}}} \\
\rowcolor{grokrow}
Grok 4.6
& 48.9 & 44.0 & 19.8 & 13.7 \\
\rowcolor{grokrow}
Composer 2.5
& 41.3 & 36.0 & 19.4 & 9.4 \\

\midrule
\rowcolor{qwenrow}
\multicolumn{5}{@{}l}{\textbf{\textit{Qwen Series}}} \\
\rowcolor{qwenrow}
Qwen3.8-Flash
& 62.3 & 54.0 & 33.8 & 19.2 \\
\rowcolor{qwenrow}
Qwen3.8-Max
& 64.0 & 58.0 & 40.6 & 18.2 \\

\midrule
\rowcolor{oursrow}
Human expert
& 92.8 & 90.0 & 77.2 & 63.4 \\

\bottomrule
\end{tabular}%
}
\end{table}






Table~\ref{tab:main-results} reports answer performance and recovery of the required video and web evidence. Among the evaluated agents, GPT-5.6 Sol achieves the highest answer accuracy (76.0\%) and the highest Web Claim recall (32.0\%), whereas Claude Opus~4.8 achieves
the highest Video State recall (43.5\%). No single evaluated agent therefore dominates all stages of the task.

These results show that answer accuracy and evidence recovery measure different aspects of task performance. For example, systems with similar answer scores can have different video- and web-evidence recall.
Because the current results provide point estimates on 100 questions, we treat these rankings as descriptive. Confidence intervals and paired tests are needed to determine whether the differences are statistically reliable.
We discuss these results further in Section~\ref{sec:discussion}.

\subsection{Qualitative Analysis}
\label{sec:error-analysis}

We further examined several incorrect predictions from GPT-5.6 Sol, which achieved the highest answer accuracy among the evaluated agents. We grouped these predictions by the earliest error that changed the subsequent research path. This analysis identified three recurring patterns. The patterns are descriptive; they do not estimate the frequency of each failure across all systems.

\paragraph{Wrong Interpretation of the Video.}
In some cases, the agent selects the wrong
entity, event, or relation from the video, but then conducts coherent web
research based on that choice. The retrieved sources may be valid and the
subsequent reasoning internally consistent, yet they support the wrong
candidate rather than the one implied by the required video evidence. As a
result, better downstream retrieval may reinforce an incorrect research path
instead of correcting it.

\paragraph{Wrong Link between the Video and the Web Search.}
In other cases, the agent identifies the relevant video content but links it to the wrong external entity or record. A shared name, visual concept, or related event can lead the search to the wrong web target, after which the remaining research may again be internally
consistent. For example, an agent may correctly identify the target object in
the video but associate it with the wrong artist or work when beginning web
search. 

\paragraph{Wrong Fact from a Relevant Page.}
In some cases, the agent reaches the correct entity and a relevant webpage but select a fact that does not match the question. This
can happen when a page contains several closely related values or records, such
as different dates, editions, events, or roles, and the agent does not use the
full question context to distinguish between them. The failure therefore occurs
after successful retrieval: the relevant source is available, but the selected
fact does not satisfy the specific constraint posed by the question.

\subsection{Discussion}
\label{sec:discussion}
\begin{table*}[t]
\centering
\caption{\textbf{Partial vs.\ complete evidence grounding.}
For each question, video states and web claims are classified as
\textbf{Complete} when all required states or claims are supported,
\textbf{Partial} when at least one but not all required states or claims are supported,
and \textbf{None} when no required state or claim is supported.
All values are percentages of the 100 benchmark questions.}
\label{tab:partial-complete}
\setlength{\tabcolsep}{7pt}
\renewcommand{\arraystretch}{1.08}
\begin{tabular}{lccc c ccc}
\toprule
& \multicolumn{3}{c}{\textbf{Video Evidence}}
&& \multicolumn{3}{c}{\textbf{Web Evidence}} \\
\cmidrule(lr){2-4}\cmidrule(lr){6-8}
\textbf{System}
& \textbf{Complete} & \textbf{Partial} & \textbf{None}
&& \textbf{Complete} & \textbf{Partial} & \textbf{None} \\
\midrule
GPT-5.6 Sol & 21 & 36 & 43 && 8 & 53 & 39 \\
Claude Opus 4.8    & 24 & 39 & 37 && 4 & 34 & 62 \\
Claude Sonnet 5  & 15 & 34 & 51 && 1 & 42 & 57 \\
GPT-5.6 Terra    & 7  & 28 & 65 && 1 & 27 & 72 \\
Grok 4.6         & 8  & 25 & 67 && 3 & 28 & 69 \\
Composer 2.5     & 7  & 27 & 66 && 1 & 23 & 76 \\
Qwen3.8-Flash     & 20 & 30 & 50 && 1 & 43 & 56 \\
Qwen3.8-Max      & \textbf{26} & 31 & 43 && 2 & 40 & 58 \\
\bottomrule
\end{tabular}
\end{table*}

\paragraph{Insight 1: Partial recovery does not readily translate into complete grounding.}
Table~\ref{tab:partial-complete} shows that current agents frequently recover
some of the required evidence without establishing the full evidence set. Complete video-evidence recovery ranges from 7\% to 26\% across the evaluated systems. Complete web-evidence recovery ranges from 1\% to 8\%. 
These results suggest that finding isolated facts is easier than establishing the full set of evidence required by a question. A joint measure of answer correctness and complete video-and-web grounding would show how often incomplete recovery affects the final answer.

\paragraph{Insight 2: Not all correct-answer paths are equally grounded.}
Answer correctness and evidence completeness do not always fail or succeed
together. Our manual inspection of correctly answered but incompletely grounded
cases reveals several distinct patterns. In some cases, the agent appears to
reach the correct answer through a reasoning path that differs from the
human-annotated one, while still establishing the same underlying semantic
dependency. The intermediate representations, ordering of steps, or source
choices may differ, yet the reasoning can converge on the same valid conclusion.
This does not imply that the required evidence is redundant; rather, human
annotators and model agents may realize the same minimal dependency through
different reasoning paths. In other cases, however, the answer is obtained by
bypassing a required modality or by directly reaching the final entity without
establishing the necessary intermediate relations. These cases should not be
treated equivalently: the former reflects a valid alternative realization of
the same evidence dependency, whereas the latter remains incomplete grounding
under the task definition.
\paragraph{Insight 3: Cross-stage constraint maintenance is a recurring challenge.}
The failure patterns in Section~\ref{sec:error-analysis} share a common structure: individual perception,
retrieval, and reasoning steps can each appear locally plausible while the
overall research trajectory is wrong. A video-side binding must continue to
constrain which web branch is valid, and question-side semantics must continue
to constrain which retrieved fact is selected. This suggests that progress in
video deep research (VDR) requires stronger mechanisms for maintaining, validating, and revising
constraints across stages, rather than improving individual stages only in
isolation. Future VDR agents may therefore benefit from explicitly tracking
intermediate entities and constraints, checking whether each new retrieval
remains consistent with earlier video evidence, and revisiting earlier decisions
when later evidence introduces a conflict.
\paragraph{Insight 4: Grounding evaluation should permit multiple valid research trajectories.}
A process-level benchmark should evaluate whether the required semantic
information has been established, rather than whether a model reproduces an
annotator's exact search sequence. \method{} therefore defines video states and
web claims at the semantic level, independently of a particular URL, wording,
or search order. This allows alternative research trajectories to receive
credit when they establish the same necessary video-to-web dependency.
\paragraph{Insight 5: Strict evidence evaluation remains challenging.}
Evaluating research processes is inherently more ambiguous than evaluating final
answers. Models may omit evidence they actually used, express an intermediate
fact at a different semantic granularity, or support the same claim through a
different valid source. Our state- and claim-level semantic evaluation accommodates variation in
annotation wording, evidence granularity, and valid retrieval paths, providing
a more robust basis for process-level evaluation than exact-match evidence
checking.
\section{Conclusion}
\label{sec:conclusion}

We introduce \method{}, an evidence-grounded benchmark for video deep
research (VDR) that explicitly connects temporally distributed video evidence with
external web evidence through structured video states, source-independent web
claims, and restricted-access shortcut checks. Evaluations of eight multimodal
agents show that strong final-answer performance does not necessarily translate
into complete evidence grounding: current systems often recover useful evidence
fragments while failing to maintain the full set of constraints across video
understanding, web retrieval, and evidence selection. Qualitative analysis
further reveals that locally plausible reasoning can still follow an incorrect
research trajectory, highlighting cross-stage constraint maintenance as an important challenge for future VDR systems. By making both outcomes and supporting
research processes auditable, \method{} provides a testbed for evaluating
this capability beyond answer accuracy alone.


\bibliography{references_fixed}
\bibliographystyle{iclr2027_conference}

\clearpage
\appendix

\section{Annotation Object}
\label{app:annotation}

The following compact schema illustrates the intended gold object. It omits implementation-only metadata.

{\small\begin{verbatim}
{
  "video": {"video_id": "...", "domain": "..."},
  "task_question": "...",
  "final_answer_gt": {"answer_text": "...", "answer_type": "entity"},
  "state_pool_gt": [{
    "state_id": "s1",
    "sub_question": "...",
    "sub_answer": "observable video fact",
    "key_evidence_units": [{
      "canonical_interval": [12.4, 16.8],
      "description": "observable fact only",
      "linked_observation_ids": ["obs_07"]
    }]
  }],
  "candidate_set": [{
    "candidate_id": "c1", "is_correct": true,
    "linked_state_ids": ["s1", "s2"]
  }],
  "web_evidence_gt": [{
    "url": "...", "claim_text": "...",
    "support_target": "candidate_disambiguation"
  }],
  "hard_negative_web": [{
    "linked_candidate_id": "c2",
    "plausibility_reason": "...", "exclusion_reason": "..."
  }]
}
\end{verbatim}}

\section{Construction Diagnostics}
\label{app:construction-diagnostics}

We report two development-stage diagnostics (Table~\ref{tab:construction-diagnostics}) to illustrate the filtering effect of
our construction protocol. These statistics refer to early development pools and
are not part of the final benchmark results.

\begin{table}[ht]
\centering
\caption{\textbf{Development-stage construction diagnostics.}
The blueprint pool and retrospective audit precede the frozen benchmark release
and are reported only to characterize the construction process.}
\label{tab:construction-diagnostics}
\setlength{\tabcolsep}{6pt}
\renewcommand{\arraystretch}{1.08}
\begin{tabular}{@{}lr@{}}
\toprule
\textbf{Diagnostic} & \textbf{Result} \\
\midrule
Candidate videos & 64 \\
Provisional question blueprints & 77 \\
Videos yielding no viable item & 23/64 (35.9\%) \\
\midrule
Earlier questions retrospectively audited & 29 \\
Passed temporal necessity & 23/29 \\
Passed retrieval necessity & 22/29 \\
Passed both necessity criteria & 22/29 \\
\bottomrule
\end{tabular}
\end{table}
\begin{table}[ht]
    \centering
    \caption{\textbf{Models, agent platforms, and reasoning settings used in the evaluation.}}
    \label{tab:model-platform-details}
    \begin{tabular}{p{0.32\linewidth} p{0.26\linewidth} p{0.24\linewidth}}
        \toprule
        \textbf{Models} & \textbf{Agent Platform} & \textbf{Reasoning Setting} \\
        \midrule
        Claude Opus 4.8; Claude Sonnet 5
            & Claude Code & High \\
        GPT-5.6 Sol; GPT-5.6 Terra
            & Codex & High \\
        Composer 2.5; Grok 4.6
            & Cursor & High \\
        Qwen3.8-Flash; Qwen3.8-Max
            & Qoder & Medium \\
        \bottomrule
    \end{tabular}
\end{table}
The early blueprint pool showed that a substantial fraction of candidate videos
did not support a natural and verifiable video deep research (VDR) question under our construction
criteria. Separately, the retrospective audit found that some structurally valid

earlier questions still failed temporal or retrieval necessity. These observations
motivated the explicit necessity checks used in the final construction pipeline.
\section{Model and Platform Details}
\label{app:model-platform-details}

Table~\ref{tab:model-platform-details} summarizes the evaluated models, agent platforms, and reasoning settings. All experiments were run on September 10, 2026. All systems received only the final system prompt described in Appendix~\ref{app:prompt}. We did not preinstall additional tools. Agents were responsible for downloading and installing any additional tools they chose to use.

We used the high reasoning setting for systems running on Claude Code, Codex, and Cursor. Qoder provides low, medium, and extra-high reasoning settings, but does not provide a high setting. We therefore used its medium setting.

\section{Evaluation Prompt}
\label{app:prompt}

To standardize the task instructions and output formats of the evaluated agents on \method{}, we employ a unified system prompt. The exact prompt provided to the models is detailed below:

\vspace{1em}
\begin{mdframed}[backgroundcolor=gray!8, linecolor=gray!35, linewidth=0.5pt, middlelinecolor=gray!35, middlelinewidth=0.5pt, roundcorner=4pt, innertopmargin=10pt, innerbottommargin=10pt, innerleftmargin=10pt, innerrightmargin=10pt, splittopskip=10pt, splitbottomskip=10pt]
\fontsize{8.5pt}{10pt}\selectfont
\begin{verbatim}
You are solving a Video Deep Research benchmark. Each item provides a 
video and a question. Use the video and any available external research
tools to answer the question.

You may choose your own research and reasoning strategy, including the 
order in which you inspect the video and search external sources.

## Data access

From each question annotation, read only:
- `sample_id`
- `task_question`
- `video.video_id`
- `video.source_uri`

Do not read or use ground-truth or annotation-only fields such as
`final_answer_gt`, `candidate_set`, `state_pool_gt`,
`web_evidence_gt`,
`hard_negative_web`, or `reasoning_checklist_gt`.

## Required output

Return:
1. the final answer;
2. the video evidence used to support it;
3. the web evidence used to support it;
4. a brief reasoning summary;
5. confidence and any limitations.

Return every answer part requested by the question in the same order.
In `final_answer.answer_text`, include only the requested final answer 
parts. Separate multiple answer parts with literal semicolons (;).
If a requested part cannot be determined, write `UNKNOWN` in that 
position.

## Video evidence

Organize video evidence as states containing one or more evidence units.

Each evidence unit must:
- refer to one continuous video interval;
- use valid start and end times in seconds;
- describe only what is directly visible or audible in that interval;
- use one modality: `visual`, `audio`, or `audiovisual`.

On-screen text counts as `visual`.

Do not invent timestamps or describe information that is notobservable 
in the specified interval.

## Web evidence

For every web source actually used, provide:
- the opened URL;
- page title when available;
- a verbatim quote from the page;
- a concise `support_claim` describing what that quote supports.

Quotes must be copied from the opened page. Do not invent URLs,titles, 
or source content.

If multiple discontinuous quote segments are used, separate them with ;.
A single verbatim segment itself must not contain ;.

## Output format

Write one JSON object per line to `vdr_predictions.jsonl`:

{
  "sample_id": "<sample id>",
  "final_answer": {
    "answer_text": "<requested final answer only>"
  },
  "used_video_evidence": [
    {
      "state_id": "s1",
      "sub_question": "<video-side question>",
      "sub_answer": "<video-side conclusion>",
      "evidence_units": [
        {
          "evidence_id": "s1_e1",
          "start_sec": 12.4,
          "end_sec": 15.1,
          "description": "<local observable fact>",
          "modality": "visual"
        }
      ]
    }
  ],
  "used_web_evidence": [
    {
      "evidence_id": "w1",
      "url": "<opened URL>",
      "title": "<page title>",
      "quote": "<verbatim quote>",
      "support_claim": "<fact supported by this quote>"
    }
  ],
  "reasoning": "<brief reasoning summary>",
  "confidence": 0.0,
  "limitations": "<uncertainties or access limitations>"
}
\end{verbatim}
\end{mdframed}

\end{document}